\documentclass[conference]{IEEEtran}
\IEEEoverridecommandlockouts

\usepackage{cite}
\usepackage{amsmath,amssymb,amsfonts}
\usepackage{algorithmic}
\usepackage{graphicx}
\usepackage{textcomp}
\usepackage{xcolor}
\usepackage{subcaption}
\usepackage{algorithm}
\usepackage{colortbl}
\usepackage{multirow}
\usepackage{dsfont}
\usepackage{hyperref}

\definecolor{LightCyan}{rgb}{0.88,1,1}

\newcommand{\ie}{\textit{i.e.}}
\newcommand{\eg}{\textit{e.g.}}

\def\BibTeX{{\rm B\kern-.05em{\sc i\kern-.025em b}\kern-.08em
    T\kern-.1667em\lower.7ex\hbox{E}\kern-.125emX}}
\begin{document}

\title{Steering Transparent Generation via Concept Bottlenecks on Energy Landscapes\\
{\thanks{This work was supported by Electronics and Telecommunications Research Institute (ETRI) grant funded by the Korean government [26ZD1120, Advancement and Commercialization of Daegu-Gyeongbuk Regional Strategic Industries(Robots, Mobility, AI, Medical, etc.)].}}}

\author{\IEEEauthorblockN{Sangwon Kim}
\IEEEauthorblockA{ETRI\\
South Korea \\
eddiekim@etri.re.kr}
\and
\IEEEauthorblockN{Kyoungoh Lee}
\IEEEauthorblockA{ETRI\\
South Korea \\
longweek7@etri.re.kr}
\and
\IEEEauthorblockN{Jeyoun Dong}
\IEEEauthorblockA{ETRI\\
South Korea \\
jydong@etri.re.kr}
\and
\IEEEauthorblockN{Kwang-Ju Kim}
\IEEEauthorblockA{ETRI\\
South Korea \\
kwangju@etri.re.kr}
}

\maketitle

\begin{abstract}

    Generative concept bottleneck models aim to enable interpretable generation by routing synthesis through explicit, user-facing concepts. In practice, prior approaches often rely on non-explicit bottleneck representations (\eg, vision cues or opaque concept embeddings) or black-box decoders to preserve image quality, which weakens the transparency. We propose \textbf{CoBELa (Concept Bottlenecks on Energy Landscapes)}, a decoder-free, energy-based framework that eliminates non-explicit bottleneck representations by conditioning generation entirely through per-concept energy functions over the latent space of a frozen pretrained generator---requiring no generator retraining and enabling post-hoc interpretation. Because these concept energies compose additively, CoBELa naturally supports compositional \emph{concept interventions}: concept conjunction and negation are realized by summing or subtracting per-concept energy terms without additional training. A diffusion-scheduled energy guidance scheme further replaces expensive MCMC chains with more stable, scheduled denoising for efficient concept-steered sampling. Experiments on CelebA-HQ and CUB-200-2011 demonstrate improvements over prior concept bottleneck generative models, achieving 75.70\%/82.42\% concept accuracy and 6.47/5.37 FID, respectively, while enabling reliable multi-concept interventions.

\end{abstract}


\section{Introduction}

\begin{figure}[t!]
    \centering
    \begin{subfigure}[b]{\linewidth}
        \centering
        \includegraphics[width=\linewidth]{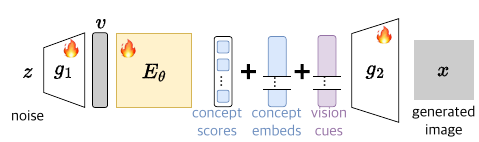}
        \caption{CBGM~\cite{cbgm}: embedding-based concept bottleneck.}
        \label{fig1:cbgm}
        \vspace{0.5em}
    \end{subfigure}

    \begin{subfigure}[b]{\linewidth}
        \centering
        \includegraphics[width=\linewidth]{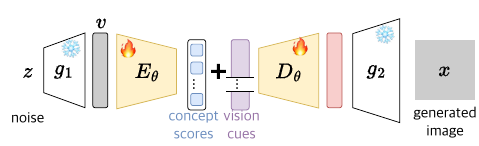}
        \caption{CB-AE~\cite{cbae}: autoencoder-based concept bottleneck.}
        \label{fig1:ae-based}
        \vspace{0.5em}
    \end{subfigure}

    \begin{subfigure}[b]{\linewidth}
        \centering
        \includegraphics[width=\linewidth]{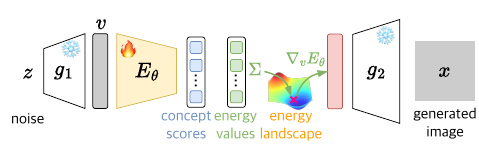}
        \caption{\textbf{CoBELa (ours)}: energy-based concept bottleneck.}
        \label{fig1:ours}
        \vspace{0.5em}
    \end{subfigure}

    \caption{\textbf{Concept bottleneck architectures for generation.} ($\boldsymbol{+}$: concat.)
        (a)~CBGM trains the generator end-to-end, concatenating concept scores with non-explicit bottleneck representations (concept embeddings and vision cues) into a single representation.
        (b)~CB-AE freezes the generator but relies on an encoder--decoder and non-explicit bottleneck representations (vision cues) that bypass the bottleneck.
        (c)~CoBELa eliminates both the decoder and non-explicit bottleneck representations; per-concept energies ($\Sigma$) and their gradient $\nabla_v E_\theta$ reconstruct the latent directly.}
    \label{fig1}
\end{figure}

Deep generative models have achieved remarkable success in synthesizing high-quality images~\cite{stygan2,adm}, enabling applications ranging from content creation to data augmentation. However, their black-box nature raises critical concerns about interpretability and intervention capability, particularly in domains where understanding \emph{why} and \emph{how} content is generated is essential---such as security-sensitive applications, medical imaging, and content moderation. The lack of transparency hinders trust, limits human oversight, and makes it difficult to debug or intervene when models produce undesired outputs.

Concept Bottleneck Models (CBMs)~\cite{cbm,cem,ECBM,eqcbm} were originally introduced to address interpretability in classification tasks by requiring models to make predictions through intermediate, human-understandable concepts. Recently, this paradigm has been extended to generative models~\cite{cbgm,cbae}, enabling semantic-level interpretation and human interventions in image generation. The core idea is to constrain the generative process to operate through an explicit concept bottleneck---a set of high-level semantic concepts (\eg, ``Male'', ``Smiling'', ``Makeup''). This makes the generation process interpretable and allows users to intervene by modifying concept values.

However, applying concept bottlenecks to generation introduces a fundamental \textbf{transparency--expressiveness trade-off}: representing high-dimensional images with a small set of discrete concepts incurs information loss that degrades visual quality. To compensate, prior generative CBMs rely on \emph{non-explicit bottleneck representations} (features that bypass the concept bottleneck without explicit semantic grounding): CBGM~\cite{cbgm} adds opaque \emph{concept embeddings} and \emph{vision cues} (Fig.~\ref{fig1:cbgm}), while CB-AE~\cite{cbae} uses \emph{vision cues} and reconstructs images through a \emph{decoder} $D_\theta$ (Fig.~\ref{fig1:ae-based}). These additions introduce hidden degrees of freedom that substantially erode the transparency of the bottleneck and blur the correspondence between explicit concepts and generated content.

To overcome these limitations, we propose \textbf{CoBELa (Concept Bottlenecks on Energy Landscapes)}, a \emph{decoder-free}, \emph{energy-based} framework that operates on a frozen pretrained generator (Fig.~\ref{fig1:ours}). CoBELa eliminates non-explicit bottleneck representations entirely and replaces the encoder--decoder with a single energy network $E_\theta$ whose concept-conditioned energies guide generation directly in the latent space of pretrained models (\eg, StyleGAN2~\cite{stygan2}). Because these concept energies compose additively~\cite{Du2020}, CoBELa naturally supports compositional interventions---concept conjunction ($c_1\wedge c_2$) and negation ($\neg c$)---by summing or subtracting the corresponding energy terms. A \emph{diffusion-scheduled energy guidance} scheme~\cite{sde} replaces expensive MCMC sampling~\cite{sgld}, yielding efficient and more stable concept-steered generation.

Our contributions are summarized as follows:

\begin{itemize}
    \item We propose \textbf{CoBELa}, a decoder-free, energy-based concept bottleneck framework that eliminates non-explicit bottleneck representations for transparent generation.

    \item We introduce diffusion-scheduled energy guidance to enable efficient and stable sampling from concept-conditioned energies without expensive MCMC.

    \item Experiments on CelebA-HQ~\cite{celeba-hq} and CUB~\cite{Wah2011} show +1.32\% / +6.86\% concept accuracy gains and $-$3.30 / $-$3.00 FID reductions over prior concept bottleneck generative models.
\end{itemize}


\section{Related Work}
\label{sec:related}

\subsection{Concept Bottleneck Models}

Concept Bottleneck Models (CBMs)~\cite{cbm,cem,ECBM,eqcbm} introduce an intermediate concept layer between inputs and predictions, enabling users to inspect and intervene on human-understandable concepts at test time. Extensions include soft concept embeddings~\cite{cem} and energy-based concept representations~\cite{ECBM, eqcbm}.

The CBM paradigm has recently been extended to generation. CBGM~\cite{cbgm} follows a CEM~\cite{cem} approach and trains the generator end-to-end, concatenating concept scores with non-explicit bottleneck representations (concept embeddings and vision cues) into a single representation (Fig.~\ref{fig1:cbgm}). CB-AE~\cite{cbae} instead freezes a pretrained generator and trains an autoencoder around the bottleneck, again augmented with non-explicit bottleneck representations (vision cues) (Fig.~\ref{fig1:ae-based}). While such non-explicit representations can recover image quality lost at the bottleneck, they compromise the transparency that the concept bottleneck is designed to provide.

\subsection{Energy-Based Models and Diffusion Guidance}

Energy-Based Models (EBMs)~\cite{lecun-ebm-tutorial} assign a scalar energy to each configuration, with lower energy indicating higher probability. A key property is \emph{compositionality}: the joint energy of multiple conditions is the sum of individual energies, enabling compositional generation by combining per-attribute energy functions additively~\cite{Du2020}. Traditional EBM sampling, however, relies on Langevin dynamics (MCMC)~\cite{sgld}, which can be expensive and unstable in high-dimensional spaces.

Diffusion models~\cite{ddpm,sde} offer a practical alternative: energy gradients can be injected into the denoising process to steer generation toward desired attributes~\cite{adm,sde,diffpure}, replacing MCMC with stable, scheduled denoising steps. Classifier-free guidance~\cite{Ho2022classifierfree} further demonstrated that conditional and unconditional score estimates can be interpolated to achieve strong steerability without external classifiers. CoBELa builds on this line of work by applying diffusion-scheduled energy guidance in the \emph{latent space} of a pretrained generator rather than in pixel space.

\subsection{Interpretable and Controllable Generation}

Interpretable generation has been approached from multiple angles. Latent space manipulation methods~\cite{Shen2020,Harkonen2020,Voynov2020} discover interpretable directions in the latent space of pretrained GANs (\eg, StyleGAN~\cite{stygan2}) corresponding to semantic attributes, enabling interventions without retraining. However, these methods often lack explicit concept grounding and may entangle multiple attributes in a single direction. Semantic conditioning approaches~\cite{Cherepkov2021,Sauer2021} train conditional models with attribute labels, but typically do not provide transparency into how concepts influence generation.

Closer to our work, recent methods aim to make pretrained generators more interpretable without modifying their weights. Network dissection~\cite{Bau2019} identifies units in GANs that correspond to semantic concepts, and activation manipulation~\cite{Goetschalckx2019} learns disentangled directions. However, these methods do not enforce a strict concept bottleneck, and the generated images still depend on uninterpretable internal representations.

\begin{figure*}[t!]
    \centering
    \includegraphics[width=\textwidth]{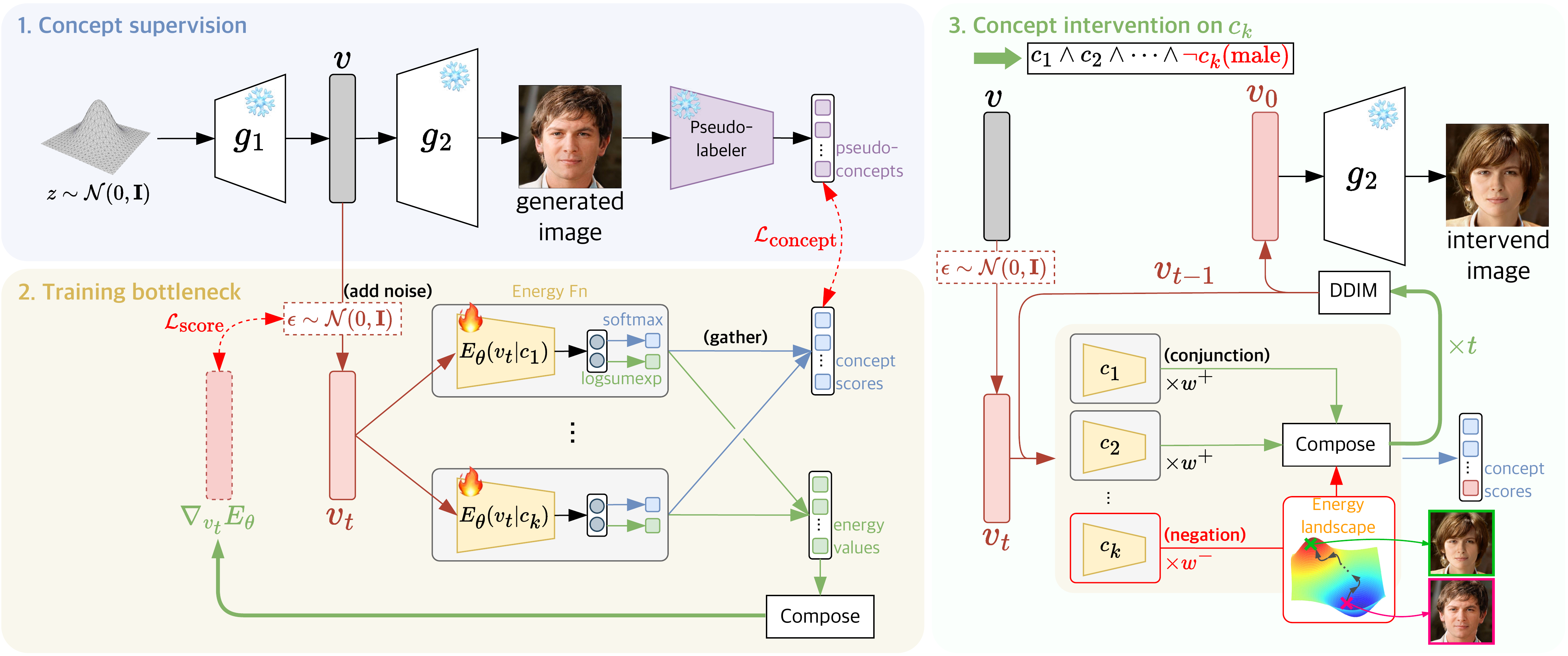}
    \caption{\textbf{Overview of the CoBELa framework.} A frozen pretrained generator is split into a mapping network $g_1$ and a synthesis network $g_2$. During training, the intermediate latent $v$ is noised and fed to the energy network $E_\theta$ along with learnable concept embeddings to produce per-concept scores (interpretable bottleneck) and energies, supervised by score-matching and concept losses. At inference, concept-weighted energy gradients guide DDIM~\cite{ddim} denoising to steer generation toward user-specified attributes.}
    \label{fig2:encobo}
\end{figure*}


\section{CoBELa: Concept Bottlenecks on Energy Landscapes}
\label{sec:method}

\subsection{Preliminaries}
\label{subsec:preliminaries}

Let $g = g_2 \circ g_1$ denote a pretrained (frozen) generator that maps noise $z \sim \mathcal{N}(0, \mathbf{I})$ to an image $x \in \mathcal{X}$. We split $g$ at an intermediate layer into a mapping network $g_1$ and a synthesis network $g_2$, writing $v = g_1(z)$ and $x = g_2(v)$ (Fig.~\ref{fig2:encobo}). Given $K$ semantic concepts, a pseudo-labeler $M$ provides binary supervision $\hat{\mathcal{S}}=M(x) \in \{0,1\}^K$ for training. We perturb the intermediate latent $v$ by sampling $t$ uniformly from $\{0,\ldots,T_s\}$ under a cosine schedule~\cite{improved-ddpm}:

\begin{equation}
    \label{eq:1_forward_noising}
    v_t                        = \sqrt{\bar{\alpha}_t}\, v + \sqrt{1-\bar{\alpha}_t}\, \epsilon,\quad \epsilon \sim \mathcal{N}(0, I).
\end{equation}

\subsection{Energy-Based Concept Bottleneck}
\label{subsec:energy}

For each concept $k$, an energy network $E_\theta$ takes $(v_t, c_k)$ as input---where $c_k \in \mathbb{R}^d$ is a learnable concept embedding looked up from a trainable embedding table---and outputs two logits:
\begin{equation}
    \label{eq:2_energy_logits}
    E_{\theta}(v_t \mid c_k) \in \mathbb{R}^2,
\end{equation}
where $E_\theta$ consists of two conditional residual blocks with FiLM conditioning~\cite{film} on concept and timestep embeddings.

The positive-class probability serves as the concept score:
\begin{equation}
    \label{eq:3_concept_score}
    s_{k} := \text{softmax}\!\left(E_{\theta}(v_t \mid c_k)\right)[1] \in [0,1],
\end{equation}
yielding a score vector $\mathcal{S}=(s_{1},\ldots,s_{K})\in[0,1]^K$ that forms the interpretable bottleneck.
We convert the logits into a scalar energy via LogSumExp~\cite{jem}:
\begin{equation}
    e_{k} := \text{LogSumExp}\!\left(E_{\theta}(v_t \mid c_k)\right).
\end{equation}

This aggregation is the log-partition function of the two-class softmax, yielding a scalar energy that is everywhere differentiable and numerically stable, while remaining in direct correspondence with the concept score in Eq.~\eqref{eq:3_concept_score}.
Per-concept energies compose additively over all $K$ concepts:
\begin{equation}
    \label{eq:4_energy_comp}
    \mathcal{E}_{\theta}(v_t) := \sum_{k=1}^{K} e_{k}.
\end{equation}

\subsection{Training Objective}
\label{subsec:training}

We train with two complementary losses. An EBM over the noised latent defines $p_t(v_t) \propto \exp(-\mathcal{E}_\theta(v_t))$, whose score satisfies $\nabla_{v_t}\log p_t(v_t) = -\nabla_{v_t}\mathcal{E}_\theta(v_t)$. Aligning this negative gradient with the added noise $\epsilon$ therefore teaches the energy landscape to assign low energy to in-distribution latent vectors, enabling $\nabla_{v_t}\mathcal{E}_\theta(v_t)$ to serve as a reliable noise predictor during sampling. Concretely, the \emph{score-matching} loss~\cite{score-matching,dsm} aligns the energy gradient with the added noise:
\begin{equation}
    \label{eq:5_score_matching}
    \mathcal{L}_{\text{score}}  = \mathbb{E}_{v,\epsilon}\left[\frac{1}{2}\left\|\epsilon - \nabla_{v_t} \mathcal{E}_{\theta}(v_t)\right\|^2\right].
\end{equation}

The \emph{concept} loss supervises per-concept logits against pseudo-labels:
\begin{equation}
    \label{eq:6_concept_loss}
    \mathcal{L}_{\text{concept}} = -\sum_{k=1}^{K} \log \text{softmax}\!\left(E_{\theta}(v_t \mid c_k)\right)\left[\hat{s}_k\right].
\end{equation}
The total objective is $\mathcal{L} = \lambda_1\, \mathcal{L}_{\text{score}} + \lambda_2\, \mathcal{L}_{\text{concept}}$.

\begin{algorithm}[t]
    \caption{Concept-Guided Sampling}
    \label{alg:sampling}
    \begin{algorithmic}[1]
        \REQUIRE Clean latent $v$, starting noise level $T_s$, intervention weights $w$, energy network $E_{\theta}$, synthesis network $g_2$, DDIM sampling steps $\mathbb{T}_{(0,T_s)}$
        \STATE $v_{t} \leftarrow \sqrt{\bar{\alpha}_{T_s}}\, v + \sqrt{1-\bar{\alpha}_{T_s}}\, \epsilon,\quad \epsilon \sim \mathcal{N}(0,I)$
        \FOR{$t$ in $\mathbb{T}_{(0,T_s)}$ from $T_s$ to $0$}
        \STATE $\hat{\epsilon}_t \leftarrow \nabla_{v_t} \sum_{k=1}^{K} w_k\, \text{LogSumExp}\!\left(E_{\theta}(v_t \mid c_k)\right)$
        \STATE $\hat{v}_0 \leftarrow (v_t - \sqrt{1-\bar{\alpha}_t}\,\hat{\epsilon}_t) / \sqrt{\bar{\alpha}_t}$
        \STATE $v_{t-1} \leftarrow \sqrt{\bar{\alpha}_{t-1}}\,\hat{v}_0 + \sqrt{1-\bar{\alpha}_{t-1}}\,\hat{\epsilon}_t$
        \ENDFOR
        \RETURN $x = g_2(v_0)$
    \end{algorithmic}
\end{algorithm}

\subsection{Concept-Guided Sampling}
\label{subsec:sampling}

At test time, each concept $k$ receives an intervention weight $w_k \in \{w^+, w^-\}$. In the absence of user edits, we set $w_k = w^+$ for all $k$, such that generation follows the learned concept energies, and human interventions are expressed by flipping the weights of selected concepts to $w^-$. This yields a weighted energy:

\begin{equation}
    \label{eq:7_weighted_energy}
    \mathcal{E}_{\theta}(v_t; w) := \sum_{k=1}^{K} w_k\, e_{k}.
\end{equation}

Setting $w_k=w^+$ for multiple concepts yields \textbf{conjunction} ($c_1 \wedge c_2$); setting $w_k=w^-$ yields \textbf{negation} ($\neg c$). We initialize the denoising trajectory from a clean latent $v=g_1(z)$ at a starting noise level $T_s$:
\begin{equation}
    \label{eq:8_init_noising}
    v_{t} = \sqrt{\bar{\alpha}_{T_s}}\, v + \sqrt{1-\bar{\alpha}_{T_s}}\, \epsilon,\quad \epsilon \sim \mathcal{N}(0,I),
\end{equation}
and denoise from $T_s$ to $0$ with a DDIM~\cite{ddim} schedule. At each step, the energy gradient serves as the predicted noise $\hat{\epsilon}_t = \nabla_{v_t} \mathcal{E}_{\theta}(v_t; w)$, and the DDIM update yields:
\begin{equation}
    \label{eq:9_ddim_update}
    \hat{v}_0 = \frac{v_t - \sqrt{1-\bar{\alpha}_{t}}\,\hat{\epsilon}_t}{\sqrt{\bar{\alpha}_{t}}},\quad
    v_{t-1} = \sqrt{\bar{\alpha}_{t-1}}\,\hat{v}_0 + \sqrt{1-\bar{\alpha}_{t-1}}\,\hat{\epsilon}_t.
\end{equation}

The final image is $x = g_2(v_0)$. The complete procedure is summarized in Algorithm~\ref{alg:sampling}.

\begin{figure*}
    \centering
    \includegraphics[width=\textwidth]{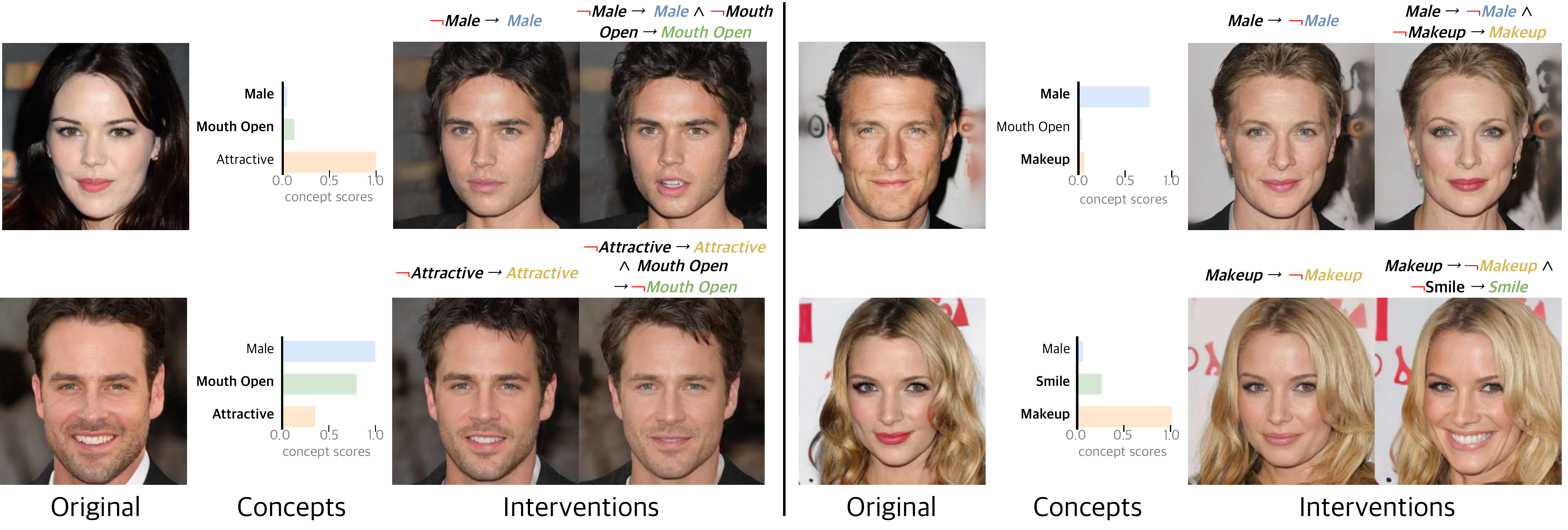}
    \caption{
        \textbf{Human-in-the-loop concept intervention on CelebA-HQ~\cite{celeba-hq}.}
        Each row shows an original generation (left), the concept scores produced by the interpretable bottleneck (bar charts), and the results of user-specified interventions (right). By default, all $K$ concepts operate at positive weight $w^+$, forming an implicit conjunction that guides generation to reflect the full concept set. The bar charts serve as explicit, human-readable explanations of the current generation: a user can inspect which concepts are active and \emph{why} the image looks as it does. To intervene, the user flips selected concepts to negative weight $w^-$ (negation, $\neg$). The key finding is that negating \emph{multiple} concepts simultaneously remains reliable: targeted attributes change as expected while non-targeted scores and facial identity are preserved---demonstrating transparent, interpretable control grounded in explicit semantic explanations.
    }
    \label{fig3:overall}
\end{figure*}

\begin{figure}[t]
    \centering
    \begin{subfigure}[b]{0.32\linewidth}
        \centering
        \includegraphics[width=\linewidth]{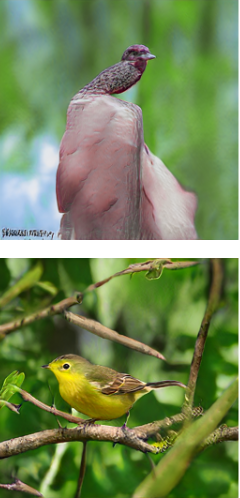}
        \caption{StyleGAN2~\cite{stygan2}}
        \label{fig4:orig}
    \end{subfigure}
    \begin{subfigure}[b]{0.32\linewidth}
        \centering
        \includegraphics[width=\linewidth]{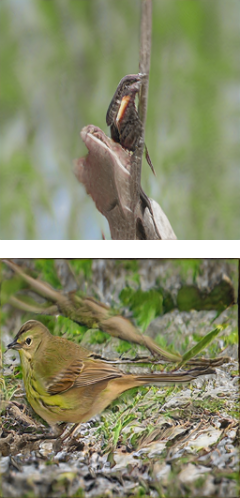}
        \caption{CB-AE~\cite{cbae}}
        \label{fig4:cbae}
    \end{subfigure}
    \begin{subfigure}[b]{0.32\linewidth}
        \centering
        \includegraphics[width=\linewidth]{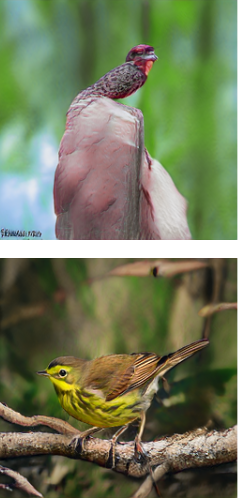}
        \caption{\textbf{CoBELa (Ours)}}
        \label{fig4:ours}
    \end{subfigure}
    \caption{\textbf{Reconstruction comparison on CUB~\cite{Wah2011}.} (a) Original generation from StyleGAN2~\cite{stygan2}, (b) reconstruction by the competing CB-AE~\cite{cbae} baseline, and (c) reconstruction by CoBELa. Our method better preserves semantic fidelity while reducing artifacts.}
    \label{fig4}
\end{figure}

\section{Experiments}
\label{sec:experiments}

We evaluate CoBELa on two benchmark datasets for concept-based generation: CelebA-HQ~\cite{celeba-hq} and CUB~\cite{Wah2011}, following the experimental protocol of prior concept bottleneck generative models such as CB-AE~\cite{cbae}. We report results on 5K samples generated from the same random seeds for both CB-AE and CoBELa for a fair comparison.

\subsection{Experimental Setup}

\noindent
\textbf{Datasets.} We use two datasets with concept annotations:
\begin{itemize}
    \item \textbf{CelebA-HQ}~\cite{celeba-hq}: $256{\times}256$ face images. Following CB-AE~\cite{cbae} for a fair comparison, we use the same $K{=}8$ most balanced concepts: \textit{smiling, male, heavy makeup, mouth open, attractive, wearing lipstick, high cheekbones, arched eyebrows}.
    \item \textbf{CUB-200-2011 (CUB)}~\cite{Wah2011}: $256{\times}256$ bird images. Following CB-AE~\cite{cbae} for a fair comparison, we use the same $K{=}10$ balanced concepts: \textit{small size, perching-like shape, solid breast pattern, black bill color, bill length shorter than head, black wing color, solid belly pattern, all-purpose bill shape, black upperparts color, white underparts color}.
\end{itemize}

\noindent
\textbf{Implementation Details.} We use pretrained StyleGAN2~\cite{stygan2} generators for both datasets. A supervised ResNet-50~\cite{He2016} classifier serves as the pseudo-labeler $M$, following the same setup as CB-AE~\cite{cbae}. We train for 50 epochs using Adam with learning rate $10^{-4}$ and a cosine noise schedule~\cite{improved-ddpm}. We set $\lambda_1=1$ and $\lambda_2=10^{-3}$, initialize interventions by noising a clean latent at $T_s=400$ (Eq.~\ref{eq:8_init_noising}), and use a $\mathbb{T}_{(0,T_s)}=50$-step DDIM~\cite{ddim} schedule. For intervention weights, we use $w^+=1$ and $w^-=-0.001$. The asymmetry is intentional. A positive weight of $1$ follows the learned energy during normal generation. By contrast, a small negative weight gently steers away from an unwanted concept without destabilizing the latent trajectory; a large negative value would overpower the energy landscape and degrade image quality.

\noindent
\textbf{Evaluation Metrics.} We measure \emph{concept accuracy} (CA), the average agreement between target and predicted concepts over all $K$ concepts. The predicted concept scores are binarized ($\mathds{1}$) at threshold $0.5$ (${}\geq 0.5 \Rightarrow 1$, ${}< 0.5 \Rightarrow 0$), then compared:
\begin{equation}
    \label{eq:10_accuracy}
    \text{CA} = \frac{1}{K} \sum_{k=1}^{K} \mathds{1}\bigl[\mathds{1}[s_k \geq 0.5] = \hat{s}_k\bigr],
\end{equation}
and image quality via Fr\'echet Inception Distance (FID; lower is better)~\cite{fid}.

\begin{table}[t]
    \centering
    \caption{Comparison with prior concept-bottleneck generative models on CelebA-HQ and CUB. $^\dagger$Number from the original paper (no reproducible code available).}
    \label{tab:main_results}
    \resizebox{\linewidth}{!}{
        \begin{tabular}{l|cc|cc}
            \hline
            \multirow{2}{*}{Method}                    & \multicolumn{2}{c|}{CelebA-HQ~\cite{celeba-hq}} & \multicolumn{2}{c}{CUB~\cite{Wah2011}}                                         \\
                                                       & CA (\%) $\uparrow$                              & FID $\downarrow$                       & CA (\%) $\uparrow$ & FID $\downarrow$ \\
            \hline
            CBGM$^\dagger$~\cite{cbgm}                 & --                                              & --                                     & --                 & 16.7             \\
            CB-AE~\cite{cbae}                          & 74.38                                           & 9.77                                   & 75.56              & 8.37             \\

            \rowcolor{lightgray}\textbf{CoBELa (Ours)} & \textbf{75.70}                                  & \textbf{6.47}                          & \textbf{82.42}     & \textbf{5.37}    \\
            \hline
        \end{tabular}}
\end{table}

\begin{table}[t]
    \centering
    \caption{Ablation of CoBELa design choices on CelebA-HQ.}
    \label{tab:ablation}
    \resizebox{\linewidth}{!}{
        \begin{tabular}{l|cc}
            \hline
            Configuration                              & CA (\%) $\uparrow$ & FID $\downarrow$ \\
            \hline
            \textbf{CoBELa (Full)}                     & \textbf{75.70}     & \textbf{6.47}    \\
            \hline
            Weak Energy Guidance ($\lambda_1=10^{-3}$) & 68.94              & 14.52            \\
            w/ MCMC (no diffusion schedule)            & 74.21              & 8.93             \\
            Fewer Diffusion Steps ($\mathbb{T}=25$)    & 74.92              & 7.01             \\
            \hline
        \end{tabular}}
\end{table}

\noindent
\textbf{Baselines.} We compare against two concept-bottleneck generative models. \textbf{CBGM}~\cite{cbgm} trains the generator end-to-end with non-explicit bottleneck representations: CEM-style~\cite{cem} concept embeddings and vision cues. \textbf{CB-AE}~\cite{cbae} applies a concept bottleneck autoencoder to a pretrained generator and also uses non-explicit bottleneck representations (\ie, vision cues). CB-AE provides reproducible code; we therefore re-run their pipeline under the same seeds and evaluation protocol as CoBELa. CBGM does not release reproducible code; we therefore report the numbers available in their paper and mark them with $^\dagger$.

\subsection{Quantitative Evaluation}

\noindent
\textbf{Baseline comparison.} As shown in Table~\ref{tab:main_results}, CoBELa improved concept accuracy over CB-AE~\cite{cbae} by +1.32\% on CelebA-HQ~\cite{celeba-hq} and +6.86\% on CUB~\cite{Wah2011}. CoBELa also reduced FID by 3.30 and 3.00 points, respectively---all \emph{without} non-explicit bottleneck representations. The only available CBGM number (CUB FID 16.7) further underscored CoBELa's advantage over embedding-based approaches, though a full comparison is precluded by the lack of reproducible code.

Non-explicit bottleneck representations, such as vision cues or opaque concept embeddings, might seem beneficial because they provide additional information at the bottleneck. However, our results suggest that these representations introduce additional degrees of freedom that may interfere with concept alignment rather than reinforce it. CoBELa instead leverages the rich latent space of a pretrained StyleGAN2 and learns concept-conditioned energies directly on this space, \emph{without} any non-explicit bottleneck features, yielding both higher concept accuracy and better FID than CB-AE.

\noindent
\textbf{Ablation.} Table~\ref{tab:ablation} isolates each design choice. Weakening energy guidance ($\lambda_1{=}10^{-3}$) caused the largest degradation (CA~68.94\%, FID~14.52), suggesting that score-matching is critical for producing energy gradients that reliably guide latent-space denoising. Replacing the diffusion schedule with Langevin MCMC~\cite{sgld} also degraded performance (CA~74.21\%, FID~8.93), consistent with the practical difficulty of stable MCMC sampling in high-dimensional latent spaces. Halving the diffusion steps to $\mathbb{T}{=}25$ yielded a modest reduction (CA~74.92\%, FID~7.01) at roughly half the inference cost, indicating a practical speed--quality trade-off for latency-sensitive scenarios.

\subsection{Concept Interventions}

Figure~\ref{fig3:overall} illustrates a human-in-the-loop workflow on CelebA-HQ~\cite{celeba-hq}. By default, all $K$ concepts are assigned positive weight $w^+$, such that the generation implicitly reflects the full concept conjunction. A user first inspects the bottleneck's concept scores to understand which concepts are active, then intervenes by flipping the weights of selected concepts to $w^-$ (negation). The bar charts confirm that the bottleneck faithfully reflects each image's concepts, providing the transparent basis for human interpretation. In single-concept negation cases---such as negating \textit{Male} or \textit{Makeup}---the targeted attribute undergoes a clear visual change while the overall facial structure and non-targeted concepts remain largely unperturbed, reflecting the localized nature of per-concept energy steering. Crucially, negating \emph{multiple} concepts simultaneously also proves reliable: flipping both $\neg$\textit{Attractive} and \textit{Mouth Open} to $w^-$ in a single denoising pass, for instance, closes the mouth and improves attractiveness without disturbing unrelated concepts such as \textit{Male}. These results suggest that CoBELa's per-concept energy landscape is sufficiently disentangled to support reliable multi-concept intervention in a transparent, human-in-the-loop setting.

\subsection{Reconstruction Quality}

Figure~\ref{fig4} compares reconstructions on CUB~\cite{Wah2011}, where fine-grained feather details provide a demanding test of bottleneck fidelity. CB-AE~\cite{cbae} reconstructions (Fig.~\ref{fig4:cbae}) exhibited visible color distortion and texture degradation. In the upper row, the gray bird lost its characteristic feather tones. In the lower row, the yellow warbler's distinctive coloring was substantially washed out. CoBELa (Fig.~\ref{fig4:ours}), operating solely through explicit concept energies without a decoder, preserved both species-specific coloring and fine surface detail more faithfully. We focused on CUB because fine-grained concepts made bottleneck-induced degradation more visually apparent than on face images, where coarse appearance could be maintained even with blurred textures. These observations are consistent with the lower FID scores reported in Table~\ref{tab:main_results}, confirming that the energy-based bottleneck provides sufficient expressiveness for high-fidelity reconstruction.


\section{Conclusion}
\label{sec:conclusion}

We introduced \textbf{CoBELa}, a decoder-free, energy-based concept bottleneck framework that steers pretrained generators solely through explicit, composable concept energies, without relying on non-explicit bottleneck representations. By replacing a decoder with concept-conditioned energy functions and diffusion-scheduled guidance, CoBELa provides transparent post-hoc interpretation over pretrained StyleGAN2 while naturally supporting conjunction and negation within a unified sampling scheme. Experiments on CelebA-HQ and CUB demonstrate gains in concept accuracy and FID over prior generative CBMs, and qualitative results show that concept-level interventions remain localized and predictable across single- and multi-concept settings.

\noindent\textbf{Limitations.}
CoBELa currently relies on a StyleGAN2 backbone; extending energy-based concept guidance to diffusion-based generators (\eg, Stable Diffusion~\cite{ldm}) is an important direction for future work. Concept supervision also depends on the quality of the pseudo-labeler $M$: classification errors in ResNet-50 propagate into the bottleneck and may reduce accuracy on visually ambiguous attributes.

    {\small
        \bibliographystyle{IEEEtran}
        \bibliography{IEEEabrv,refs}
    }

\end{document}